# Fashion-model pose recommendation and generation using machine learning

Vijitha kannumuru(Vijitha.kanumuru@gmail.com) , Santhosh Kannan S P (mail2santhoshkannan@gmail.com ), Krithiga Shankar (skrithi05@gmail.com), Joy Larnyoh (joylarnyoh@gmail.com ), Rohith Mahadevan (rohithmahadev30@gmail.com), Raja CSP Raman (raja.csp@gmail.com)

**Abstract**

Fashion-model pose is an important attribute in the fashion industry. Creative directors, modeling production houses and top photographers always look for professional models able to pose. without the skill to correctly pose, their chances of landing professional modeling employment are regrettably quite little. There are occasions when models and photographers are unsure of the best pose to strike while taking photographs. This research concentrates on suggesting the fashion personnels a series of similar images based on the input image. The image is segmented into different parts and similar images are suggested for the user. This was achieved by calculating the color histogram of the input image and applying the same for all the images in the dataset and comparing the histograms. Synthetic images have become popular to avoid privacy concerns and to overcome high cost of photoshoots. Hence, this paper also extends the work of generating synthetic images from the recommendation engine using styleGAN to an extent.

## 1. Introduction:

Fashion photography is often remarked as an alluring and high-paying career that requires sufficient hard work. However, with a proper understanding of its history and pursuing some tips, one can become an eminent fashion photographer. It falls under the domain of photography genres, commercial photography. It has become a big part of the art world with its growing popularity. Fashion photography is ubiquitous, and one can see a fair amount of images of men and women dressed, showcasing outfits, accessories, and footwear. The use of machine learning in the fashion industry is very minimal. Thus, not many fashion photographers use machine learning technologies for their existence.

Machine Learning is the ability to make the computer learn by training it with a sample of data and not by explicit conditional programming. It is basically a subset of AI. ML models perform poorly when it comes to unstructured data (image / video) which forces us to choose deep learning techniques. Deep Learning is a machine learning technique that uses neural network algorithms that mimics a human brain called neurons, allowing it to learn from large amounts of data. A neural network architecture can have any number of hidden layers with one input and one output layer. Several real time applications like autonomous vehicles that do object detection and chatbots used by many companies for better customer experience are well integrated into our day to day lives.

Generative adversarial networks(GANs) is one of the top neural network techniques that is broadly used in the AI field. GAN's architecture involves two sub-models: a generator, for generating new examples and a discriminator, for classifying whether generated examples are real or fake. One of the best use cases of GANs is the creation of synthetic images which are pictures generated using computer graphics & simulation methods of Artificial Intelligence (AI), to represent reality with high fidelity.

Due to high unavailability of public data sets & cost of data collection, synthetic image creation has become a major role player in creating data for numerous real time scenarios like fault detection in production lines, creating images of models at different locations and many others. This helps with the objective of "BC" this paper to produce synthetic images with high data quality, scalability, accuracy with less noise.

## 2. Literature survey:

Photography is a "science", an "art" and a "commerce" either and it is used in many fields such as landscapes, food, portraits, people, the announcement of a product and fashion. And where we have another science which could be related to photography which is called "science of fashion" to issue the science of "fashion photography". And in this research we will go over this science to detect its aspects and standards in order to help the fields which depend on it such as brand-names, brand stores and Officials ad campaigns. [11]

Martínez et al [12] in their work suggested that it is important that the photographer manages to capture part of the essence of the style that he wants to coordinate. It is for this reason that the issue of processes within fashion photography acquires a very relevant role, perhaps even more important than the final result, since, within those moments, it is where identity begins to develop, the creativity and identity of the photographer, of the model and therefore of the fashion that it is intended to reflect, and later coordinate for a future assimilation with the viewer of the innovative and dynamic art of fashion photography.

CNN has introduced several models for image generation and translation to improve the modeling of nonlinear mapping from input to output and producing more realistic images The most prominent models among them are the GAN models [2] .There are various versions of the GAN algorithm for different applications. For our study, Vanilla GAN, StyleGAN, and cGAN based versions were examined [1]

The GAN model needs to be fed with training datasets, which could either be paired or unpaired. Dealing with unpaired training datasets is difficult, because mapping between input and output does not exist. [2]. We have to deal with unpaired datasets in our research with no information provided as to which input matches which output.

CycleGAN is a very popular GAN architecture primarily being used to learn transformation between images of different styles like mapping between realistic and artistic image or transformation between images of horse and Zebra or a transformation between winter image and summer image. [3] Style-GAN2 generator starts from a learned constant input and adjusts the "style" of the image at each convolution layer based on the latent code, therefore directly controlling the strength of image features at different scales [4].

The similar training techniques can be employed into other state-of-the-art GAN architectures in spite of the other two mentioned previously and evaluate them using standard image quality metrics and we can show the generation of large sets of high-fidelity synthetic images using a very small set of training data[8].

The target image generated GAN architecture has to be acceptable and hence the images should have a very good quality along with the required effects. Hence Synthetic images are generated with only the source light used without considering lighting controls like direction and concentration and this affects the effects that can be achieved. A directed light can be modeled mathematically as the light emitted by a single point specular reflecting surface illuminated by a hypothetical point source light. Another light control is a variable sized cone surrounding the light direction. This feature is not found on studio lights, but it can be used to produce a sharply delineated spotlight (as opposed to the tapering off which occurs using the concentration exponent) or to isolate a light on a single object in a scene composed of several objects [9].

Training machine learning models on standard synthetic images is problematic as the images may not be realistic enough, leading the model to learn details present only in synthetic images and failing to generalize well on real images. The goal of 'improving realism' is to make the images look as realistic as possible to improve the test accuracy. These researchers created a deep neural network, called 'refiner network', that processes synthetic images to improve their realism. They used an auxiliary discriminator network that classifies the real and the refined (or fake) images into two classes. The refiner network tries to fool this discriminator network into thinking the refined images are the real ones. Both the networks train alternately, and training stops when the discriminator cannot distinguish the real images from the fake ones. The subjects found it very difficult to tell the difference between the real and refined images. THey evaluated the study and 10 subjects chose the correct label 517 times out of 1000 trials, meaning they were not able to reliably distinguish real images from refined synthetic ones. [10]

## 3. Working Methodologies

Recommender systems are well established concepts which are basically the algorithm that recommends items to the users based on certain criteria similarity or similar patterns found in the given data. The recommenders are of two types, Content-based recommender : depends on the similarities found among the data, Collaborative-filtering : relies on the user's preference similarities and tracking one user activities based on another user for recommendation.

For this research, we don't require any user based behaviors so we are using the Content-based recommender concept. This would be the base step from which the whole architecture is built. Firstly, the recommender app would group the input images based on the color and resolution and suggest a set of images as output which will form the foundation for the next stage.

The first part of the research is an exploration of similar poses. The model has been trained with a training dataset consisting of pictures containing poses. This enables the model to be able to suggest similar poses when fed with an image. The next part is the Generative Adversarial Network (GAN) model, a type of unsupervised ML algorithm that creates new data points based on the existing data with the help of a generative neural network (learns to generate similar data points from existing) and discriminative neural network (learns to distinguish b/w real & generated data points) will be used in order to attain the objective of this research.

As indicated previously, the recommender API's output now serves as input for the GAN to generate synthetics images with different poses. We have several GANs available for creating synthetic images, such as Projected GAN, Fast GAN, Style GAN and many. For the use case, Style GAN is considered as a powerful architecture when it comes to creating more realistic and high resolution facial images.

## 4. Architecture of the Model:

The styleGAN architecture being used in this research needs as low as only 12 images for training. Hence to improve the image quality of the training, an image recommender program is developed which will take an input image and outputs 12 similar images to the input. The similarity of the images will be based on the color and histogram score. With this approach we will be selecting similar images from the range of 1500 images, to train the GAN architecture which will improve the fake face generation technique.

Fig 6.1.1 shows the recommender architecture and fig 6.1.2 shows the style GAN architecture.

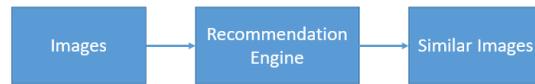

**Fig 4.1**

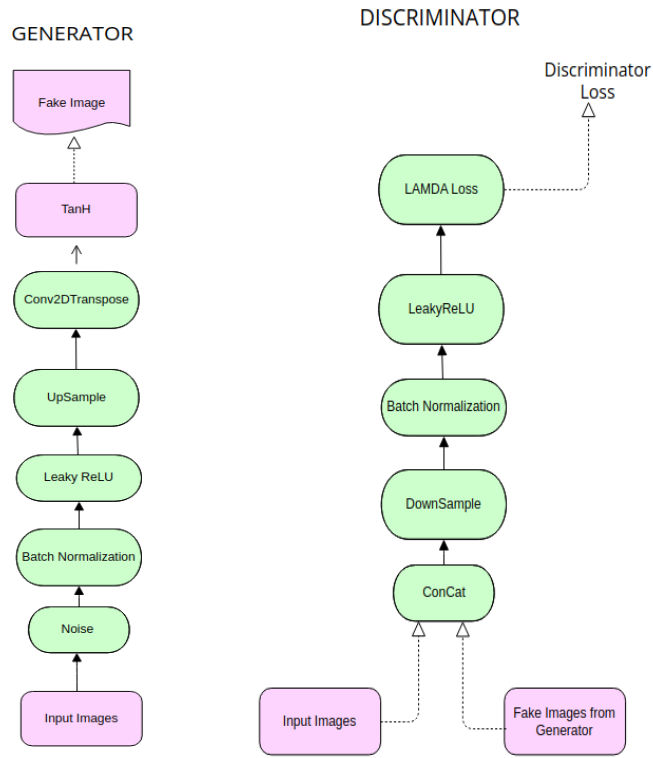

**Fig 4.2**

## 5.1 Data collection method

Data collection method typically involves either primary or secondary data collection. We have used a primary data collection method for our research in order to get a focused image which will be suitable for our scenario rather than having a generalized secondary data collection.

The objective of this research focuses on generating synthesized mugshots or headshots, so we have gathered only facial image data by using web scrapers to collect images from search engines, websites and social media platforms like instagram & Pinterest. Also have employed the Python OpenCV library to read and manipulate the images. Other options were explored as well like Python Imaging Library(PIL) and Python Requests library for gathering images, but decided on the approach above mentioned as it was adequate for the needs.

The model training is done next. The Generator is fed with the two datasets and the noise generated. The output from the generator will be input to the discriminator along with the training data set. Hence the final synthetic image of a human face is generated.

**5.2 Working of the Model**

Pose recommender API will be provided with an input image that will generate 12 similar images as output and it will simply take an input to generate similar images based on the pixels and the color of the input image. These generated images will be the input to the GAN architecture.

For this research, StyleGAN architecture was used to generate the synthetic images. The output from the recommender application will be split into two different datasets in which each data set contains 6 different images. Hence the predicted image will be based on the cross product of these two input datasets where 6*6 iterations will happen for any given epoch size. Here, 2000 epochs have been considered.

Downsampling in the discriminator network was then accomplished to reduce the spatial resolution of the image using convolutional layers. Followed by upsampling operation using transposed convolutional layers in the generator network to increase the spatial resolution of the images.

Binary cross entropy and Adam optimizer was used for generator and discriminator losses and optimizations, respectively, to be defined.

**6. Use case of the work**

The research was applied to the fashion industry, specifically to the models who take pictures for fashion brands. Most of them have a tough time getting the poses right when they are to take pictures. With the output of this research work , they are given recommended poses that would give them for subsequent picture taking. All they need to do is to feed in an image of a pose and the model would generate poses similar to that. It is also relevant to the photographer in that it gives an idea of how the model may pose and how to properly do the capturing. The photographer get to know the light intensity, the photo angle, the kind of lens and other technologies to replicate the quality of the pictures in the pose recommenders

The advantage of using this technique is to have heterogeneity in the images for better learning of the model. The system is flexible, easily scalable, and produces complex and diverse images with pixel-perfect annotations at no cost. Generative Adversarial Networks (GANs) are also used to generate images for small data scenarios augmentation Further, it is used in Advertising and marketing. They are used to create product images as well as generate influencer promotion images.

**7. Result analysis:**

A recommender application is used to generate similar poses. This was achieved by calculating the color histogram of the input image and applying the same for all the images in the dataset and comparing the histograms. The histograms are calculated by using the CV2 library. Various algorithms were tested for comparing the similar images including - Correlation, Chi-Squared, Intersection and Bhattacharyya. We concluded on using the Bhattacharyya method for comparison.

Below is the process to perform the color histogram:
1. Get input image
2. Calculate color histogram of the input image
3. Calculate color histogram for the other images in the dataset

4. Compare the input image histogram with that of the other images in the dataset.
5. Sort the images in descending order to get the more similar images
6. Saved the top 12 images in an output folder

Below is the snippet of the results generated using the recommender engine.

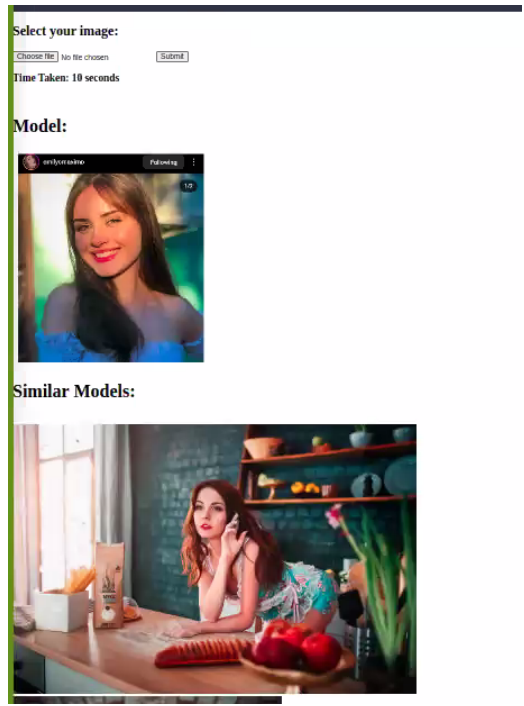

**Fig 7.1**

The output images from the recommender engine are fed to the StyleGAN architecture to generate synthetic images. The StyleGAN architecture requires only 12 images to generate a synthetic image which is very promising as most of the GAN's require a lot more training size from 30K to 200K Images to generate proper images.

**i) Simple GAN:**
A generative model was made using simple GAN. The model ran for many 6 generations with 150 epochs and 2000 input images. It generated a sample facial image as shown in fig 7.1.1 The images are not properly formed as the architecture had reached the converging threshold point.

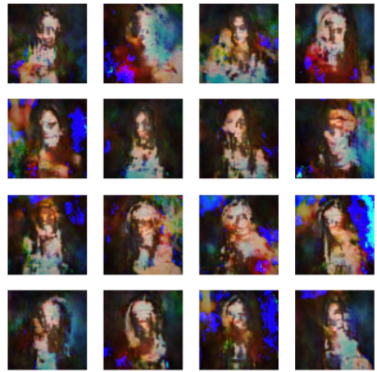

**Fig 7.2**

### ii) Style GAN :

Style GAN (Karras et al. 2019) is one of the GAN architectures that stands out more than others by providing remarkably well behaved latent space along with high resolution synthetic images without pretrained feature space. It can be accomplished by converting the input noise vector into an "intermediate latent code" via a mapping network. These can be used to generate a variety of "styles" with various levels of detail, from the broad strokes to the intricate nuances of an image with a very small number of input dataset.

The effectiveness of the model performance can be calculated using various computational metrics like the input data size, number of epochs, time taken to execute the model, batch size and data resolution.

In this research we have executed the Style GAN multiple times by altering the metrics with different parameters. As a result, it is visible in Fig 7.3, 7.4 and 7.5 which shows the visual comparison of the real facial images and the synthetic images that are generated using the mentioned architecture.

Fig 7.3 is the result obtained for 12 input images with 1000 epoch size where noise vectors are layered on top of the source images with 20 minutes of computational time.

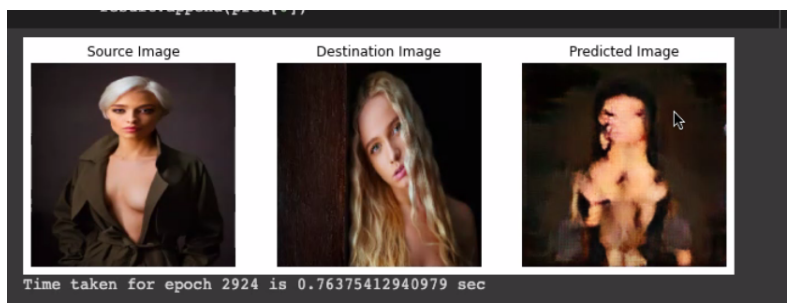

**Fig 7.3**

Fig 7.4 below is the output generated by the style GAN that fed on 12 input images from the recommendation engine and executed for 2000 epoch size shows that the synthetic generated has improved based on the increased number of iterations.

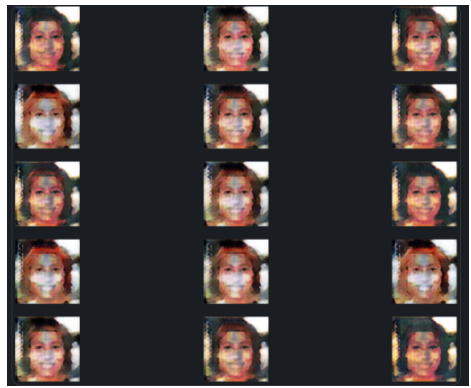
**Fig 7.4**

## 8. Future Work:

Currently, this work generates a picture of a human face using StyleGAN Architecture. The scope of the work can be increased to generate a synthetic Image with higher resolution and quality. This can be further extended to create a synthetic image of the full body of a human by creating synthetic images for different poses that can be used in the retail industry for showcasing/advertising the product on commercial websites. This could save time and money spent on hiring a model and carrying out photoshoots.

By introducing other concepts like Transmission mapping, Atmospheric Lighting or Topology to the existing GAN architecture, it is possible to generate more texture-rich & high-density images that can be helpful in training medical diagnosis algorithms where the data generated can be used to detect diseases or anomalies more precisely or In facial recognition, the data generated can be used to train algorithms to recognize different facial features and expressions.

## 9. Conclusion :

Pose recommender basically provides personalized suggestions of poses and angles for fashion models and photographers in order to develop their own signature poses and angles that showcase their unique look. By providing personalized recommendations, the current pose recommender in this research helps to group the most flattering, unique, and impactful images for their portfolios based on the sample input given. This allows photographers to convey the style and look of the brand they are shooting for.

The current approach can be used to create realistic facial images models for characters in the gaming industry, as well as for facial animation in virtual reality and augmented reality applications which was highly time consuming and required a lot of computational power.

Due to the highly complex nature of the architecture, it ended up generating images in disarray containing distortion and lack of definition in the first few trials, but it was solved by fine tuning the hyperparameter along with increasing the epochs sizes Furthermore, it can generate realistic facial expressions for characters in a movie or TV show that helps with realism of the characters also, it can be utilized to produce more expressive facial animation for virtual avatars and virtual characters.

## 10. Limitations of the research:

Even though the model is running successfully by recommending similar poses, the generation of poses cannot be achieved at a 100 percent accuracy rate. As a first part of the research only facial images are being generated by the system with the output from the recommender engine. The future work is to improve the output of the system and generate accurate results.